\documentclass[runningheads,a4paper]{llncs}

\usepackage{amssymb}
\usepackage{graphicx}

\usepackage{url}
  
\newcommand{\keywords}[1]{\par\addvspace\baselineskip
\noindent\keywordname\enspace\ignorespaces#1}

\usepackage[super]{nth}
\usepackage{algorithm}
\usepackage[noend]{algorithmic}

\usepackage{amsmath}
\usepackage{mathtools}

\DeclareMathOperator*{\argmax}{argmax}
\DeclareMathOperator*{\argmin}{argmin}

\begin{document}

\mainmatter  

\title{Data Driven Aircraft Trajectory Prediction with Deep Imitation Learning}

\titlerunning{Data Driven Aircraft Trajectory Prediction with Deep Imitation Learning}

%
%
\author{Alevizos Bastas%
\and Theocharis Kravaris \and George A. Vouros}
\authorrunning{A. Bastas, T.Kravaris, G.A.Vouros}

\institute{AI Lab, Department of Digital Systems, University of Piraeus, Greece\\
\url{http://ai-group.ds.unipi.gr/ai-group/}}

%
%
%
\maketitle

\begin{abstract}

The current Air Traffic Management (ATM) system worldwide has reached its limits in terms of predictability, efficiency and cost effectiveness. Different initiatives worldwide propose trajectory-oriented transformations that require high fidelity aircraft trajectory planning and prediction capabilities, supporting the trajectory life cycle at all stages efficiently. Recently proposed data-driven trajectory prediction approaches provide promising results. In this paper we approach the data-driven trajectory prediction problem as an imitation learning task, where we aim to imitate experts "shaping" the trajectory. Towards this goal we present a comprehensive framework comprising the Generative Adversarial Imitation Learning state of the art method, in a pipeline with trajectory clustering and classification methods. This approach, compared to  other approaches, can provide accurate predictions for the whole trajectory (i.e. with a prediction horizon until reaching the destination) both at the pre-tactical (i.e. starting at the departure airport at a specific time instant) and at the tactical (i.e. from any state while flying) stages, compared to state of the art approaches. 

\end{abstract}


\keywords{Imitation Learning, Inverse Reinforcement Learning, Trajectory Prediction, Aircraft Trajectory}



\section{Introduction}
Nowadays, the ATM system is based on an airspace management paradigm that leads to demand imbalances that cannot be dynamically adjusted. This entails, among others, higher air traffic controllers’ (ATCO) workload, which determines the maximum system capacity. With the aim of overcoming ATM system drawbacks, different initiatives, dominated by Single European Sky ATM Research (SESAR) in Europe and Next Gen in the US, have promoted the transformation of the current environment towards a new trajectory based ATM paradigm. In this Trajectory Based Operations paradigm the trajectory becomes the cornerstone upon which all the ATM capabilities will rely on, supporting the whole trajectory life cycle: From the trajectory planning, to the negotiation and agreement, execution, amendment and modification stages.  

The proposed transformation requires high fidelity aircraft trajectory planning and prediction capabilities, supporting the trajectory life cycle at all stages efficiently. Indeed, predictability is considered as the main driver to enhance operational key performance areas, such as capacity, efficiency for all stakeholders (i.e. airspace users, air traffic controllers, network manager, airspace navigation service providers, etc), and, of course, safety. Enhancing predictability and bringing more automation into all stages of operations  emerges both, as a mid-term need (the European Network Manager EUROCONTROL, forecasts increases in traffic of +50\% in 2035 compared to 2017, meaning 16 million flights across Europe) and as a long-term need (2035+).

Current trajectory predictors are based on deterministic formulations of the aircraft motion problem. Although there are sophisticated solutions that reach high levels of accuracy, all approaches are intrinsic simplifications to the actual aircraft behavior, which delivers appropriate results for a reasonable computational cost. Predictors' outputs are generated based on apriori knowledge of the  flight plan (i.e. airline's planned and intended trajectory filed before departure), the expected command and control strategies released by the pilot, or the Flight Management System (FMS) instructions (known as Aircraft Intent \cite{b1}), a forecast of weather conditions to be faced throughout the trajectory, and the aircraft performance. These model-based  or physics-based approaches are deterministic: They return always the same trajectory prediction for a set of identical inputs. 
Although the use of the concept of Aircraft Intent together with very precise aircraft performance models such as  Base of Aircraft Data (BADA) \cite{b2} has helped to improve the prediction accuracy, the model based approach requires a set of input data that typically are not precisely known (i.e. initial aircraft weight, pilot/FMS flight modes, etc.). In addition, accuracy varies depending on the intended prediction horizon (look-ahead time).

Recent efforts in the field of aircraft trajectory prediction have explored the application of statistical analysis and machine learning techniques to capture non-deterministic influences for aircraft trajectory prediction. Linear regression models \cite{b4} \cite{Hamed2013StatisticalPO} and neural networks \cite{b5} \cite{Cheng2003DataMF} \cite{liu2018predicting}, have returned successful outcomes for improving the trajectory prediction accuracy for traffic flow forecasting. Generalized Linear Models \cite{doi:10.2514/6.2013-4782} have been applied for the trajectory prediction in arrival management scenarios and multiple linear regression \cite{b8} \cite{b99} for predicting estimated times of arrival (ETA). These efforts include as input dataset historical surveillance data, and additional supporting data required for robust and reliable trajectory predictions (e.g. flight plans, airspace structure, Air Traffic Control procedures, airline strategy, weather forecasts, etc.) depending on their objectives. However, these approaches make specific assumptions, are restricted to a specific operational/tactical phase, have a limited prediction horizon, or consider specific constraints for trajectories.

In this paper we approach the trajectory prediction problem as a data-driven imitation learning problem, where we aim to imitate the experts ``shaping'' the trajectory, learning models that incorporate their preferences, strategies, practices etc.\footnote{Subsequently preferences, strategies, practice etc. are termed as ``policy".} in an aggregated way. Towards this goal we present a comprehensive framework that comprises the Generative Adversarial Imitation Learning state of the art  method, in a pipeline with trajectory clustering and classification methods. This approach can be effective (in terms of accuracy of predictions) even with a small number of historical trajectories, compared to  other approaches, and more importantly, can provide accurate long-term predictions both at the pre-tactical (i.e. before departure, given the departure airport and the take-off time) and at the tactical (i.e. while flying, given a state en-route) stages, compared to state of the art approaches. We provide a series of experiments, using demonstrated trajectories from Barcelona to Madrid, showing the effectiveness of our approach.

Major contributions that this paper makes are as follows:
\begin{itemize}
    \item It formalizes the trajectory prediction problem as an imitation learning process, given a set of historical trajectories provided as ``expert" demonstrations, thus considering an aggregation of the individual stakeholders' policies "shaping" these trajectories.
    \item It introduces a framework that is able to detect different classes (patterns) of trajectories, learns models to identify the most likely class to which a future trajectory belongs exploiting forecast contextual features (weather conditions), and applies stochastic policies to predict trajectories, subject to their membership in a specific class.
    \item It proposes the straightforward use of state of the art deep imitation learning methods, which are able to learn trajectory models without making any assumption on the form of a cost function, in continuous state-action spaces and with no specific requirements on specifying trajectory constraints; and finally, with minimal data pre-processing requirements.
    \item It provides experimental results that - although they concern a specific origin-destination airports pair - show the far in the future prediction abilities of the method, either at the pre-tactical or at the tactical stage of operations.
\end{itemize}
The structure of the paper is as follows: Section 2 provides background knowledge and defines raw and enriched trajectories, it specifies the trajectory prediction problem in its more generic form, and provides background knowledge on imitation learning. Section 3 formalizes the data-driven aircraft trajectory prediction problem as an imitation learning process, Section 4 presents the overall prediction framework with emphasis on the Generative Adversarial Imitation Learning method used, and Section 5 presents experimental results. Finally, Section 6 presents related work and discusses the effectiveness of the proposed method w.r.t. state of the art trajectory prediction methods in the aviation domain. Section 7 concludes the paper.


\section{Background Knowledge}

\subsection{Aircraft Trajectory Prediction}
\label{sec:prediction}

In the aviation domain, a trajectory is defined as the description of movement of an aircraft both in the air and on the ground.\footnote{https://ext.eurocontrol.int/lexicon/index.php/Trajectory}. 
This description can be provided by a chronologically ordered sequence
of aircraft states described by a list of state variables. Most relevant state variables are airspeeds, 3D position (determined by latitude (f), longitude (l) and geodetic altitude (h)), the bearing (c) or heading (y) and the instantaneous aircraft mass (m). 
Trajectories providing only spatio-temporal information at each state (i.e. 3D positions and timestamps)  can be detected by exploiting surveillance data and are called {\em raw trajectories}.  

More formally, a {\em raw trajectory} $T$ of an aircraft is defined to be a sequence of $|T|$ pairs $s_i$=$<p_i, t_i>$, $i\in [0,|T|-1$, where $p_i$ is a point ($l,f,h$) in the 3D space (position) 
and $t_i$ is a timestamp. In this case a trajectory state is represented by a 4D point, and the $length$ of a trajectory $T$ is equal to the number of states $|T|$. 

Following a data-driven approach, we aim to exploit historical 4D aircraft trajectories whose states include 3D aircraft position with timestamps, in conjunction to other contextual features that may provide useful features in the prediction process, such as weather conditions at each state, traffic, special events occurring etc.
Adding variables in a trajectory state, results in a trajectory with {\em enriched points} or {\em enriched states}, thus to an {\em enriched trajectory}:

An {\em enriched  trajectory state} or {\em enriched  trajectory point}, corresponding to a raw trajectory point $s_i$=$<p_i, t_i>$ is defined to be a triplet $s_{r,i}$=$<p_i, t_i, v_i>$, where $v_i$ is a vector consisting of categorical and/or numerical variables annotating the raw trajectory state. An {\em enriched trajectory} $T$ is defined to be a sequence of enriched states $s_{r,i}$=$<p_i, t_i, v_i>$, $i\in [0,|T|-1]$.


In the aviation domain, towards implementing the Trajectory Based Operations paradigm, predictability of trajectories is of immense importance. Indeed, uncertainties occurring during a flight have impact on multiple stakeholders, including airspace users (i.e. airlines), the air navigation service providers (ANSPs) providing services for Air Traffic Management, the air traffic controllers, as well as ground operators and of course, passengers. Confronting uncertainties and adopting to them is costly for all. For instance, these may require assigning  delays to flights, or choosing alternative routes to those planned for a flight, resulting to more fuel consumption, more workload for Air Traffic Controllers (challenging the capacity of the ATM system), and cascaded effects to the whole ATM system.

A {\em predicted trajectory} can be defined as the future evolution of the
aircraft state as a function of (a) the current flight conditions (e.g. an initial state with actual weather conditions), (b) a forecast of contextual features (e.g. forecast weather
conditions at specific positions) and (c) a description of how the aircraft is to transit among subsequent states starting from an initial state and on, i.e. a ``policy" on how the trajectory evolves. 

Casting the trajectory prediction to a data-driven problem, and assuming a set $\textbf{T}_E=\{T_{E,i}|i=1,2,3,...\}$ of historical, demonstrated enriched trajectories, the {\em trajectory prediction problem} can be defined as follows:
Given ${\bf T}_E$ and a ``cost"\footnote{In the context of data driven methods, the cost function denotes a penalty for low adherence to demonstrated data.} function $c$, the objective is to predict a trajectory $T_\pi$ such that 

\begin{equation}
T_\pi=\argmin_\pi\ \mathbb{E}_\pi[c(\langle p,t,v \rangle,a)]\  \label{eq:pred}
\end{equation}

\noindent where $\mathbb{E}$ denotes the expected cumulative costs for all states $s$ generated  along the trajectory by following a policy $\pi(a|s)$ prescribing the probability of applying an action $a$ at an enriched state $s=\langle p,t,v \rangle$, (we discuss about states and actions in a subsequent paragraph). Actually, according to equation \ref{eq:pred}, the ultimate objective is to find the policy $\pi$ that determines the generation of a minimal-expected-cumulative-cost predicted trajectory $T_\pi$.

The cost function may take several forms depending on how the problem is approached: For instance, considering specific trajectories (e.g. flight plans, or cluster medoids) as constraints to which the predicted trajectory must adhere to (e.g. as in \cite{Georgiou2019SemanticawareAT}), and measuring the adherence of predictions to these constraints, the cost function may take the form of a distance function between these trajectories and predicted trajectories. Other constraints may also be incorporated into the process, such as those depending on the aircraft type, allowable or desired states, origin and destination airports, etc. Generally, in a data-driven trajectory prediction process, the cost function shows the adherence of predictions to patterns, constraints and policies regarding historical cases. We delve into this issue further while formulating the trajectory prediction problem as an imitation process, in Section \ref{sec:formulation}.

A final note on equation \ref{eq:pred}, concerns the trajectory states, additional features and actions considered: The formulation indicates separately the 4D position information with timestamps and other variables enriching states. Indeed, additional features may be considered in the cost function, such as weather variables, traffic, airspaces crossed, etc. Also, different prediction processes may have different prediction objectives: For instance, one may predict the aircraft position at specific time instances, or predict the time instance that a specific position will be reached, or the position together with the corresponding timestamp, or even predict some of the contextual features, such as aispaces crossed at specific time instances, or traffic. What we aim to predict in this work is the 3D aircraft position at specific time instances.  Actions executed at each state determine how the trajectory evolves towards the next state (e.g. by means of change in speed, change in aircraft direction (bearing), other detailed aircraft intent instructions etc.). Actions may also vary between different approaches. Let $A$ be the set of actions assumed.


\subsection{Imitation Learning}
\label{sec:imitation}
Making assumptions on, or handcrafting the cost function is crucial to the prediction process, as flown trajectories are shaped by several stakeholders each with own preferences, strategies and concerns. Thus, we are motivated to apply an imitation learning approach towards learning a policy modeling the way stakeholders shape the evolution of trajectories, considering historical trajectories as experts' demonstrations. 

Imitation learning studies the problem of learning to perform a specific task in a setting, where the learner has access to expert demonstrations, but cannot query the expert for more samples while being trained, and has no access to a cost signal. There are two fundamental approaches to imitation learning: Behavioral Cloning (BC) and Inverse Reinforcement Learning (IRL).

Behavioral Cloning \cite{pomerleau1991efficient} addresses the imitation learning problem as a supervised learning problem over the state-action pairs of the expert demonstrations. Behavioral Cloning solves a regression problem minimizing the error between the actions demonstrated and the policy actions, over the states of the historical trajectories. This technique suffers from compounding errors and a regret bound that grows quadratically in the time horizon of the task leading to poor performance \cite{ross2010efficient,ross2011reduction}. 

Inverse Reinforcement Learning \cite{abbeel2004apprenticeship,ziebart2008maximum,finn2016guided,finn2016connection} on the other hand, aims at deriving a cost function that assigns minimal cost to trajectories demonstrated by experts and maximal cost to trajectories generated by other policies. Given that many policies may demonstrate the same trajectories, the maximum entropy inverse reinforcement learning approach aims to find the maximum entropy policy \cite{ziebart2008maximum}. 
Actually, this process comprises two steps: The first one outputs a desired cost function according to the following formula,

\begin{equation}
    IRL_\psi(\pi_E)=\argmax_{c \in \mathbb{R}^{S \times A}}-\psi(c)+(\min_{\pi \in \Pi}-\lambda_HH(\pi)+\mathbb{E}_\pi[c(s,a)])-\mathbb{E}_{\pi_E}[c(s,a)] \label{eq:maxent}
\end{equation}

\noindent where, $\psi(c):\mathbb{R}^{S \times A} \rightarrow \mathbb{R} \cup \{\infty\}$ is a convex cost function regularizer, $\pi_E$ is the expert policy (provided by the demonstrated trajectories) and $\Pi$ is the set of all policies,  $H(\pi)$ the entropy function and $\lambda_H$ its weighting parameter. The second step is to find a policy that minimizes the expected cumulative cost and maximizes the entropy by using the cost function into a standard reinforcement learning problem, very close to the one specified by equation (\ref{eq:pred}):

\begin{equation}
    RL(c)=\argmin_{\pi \in \Pi} -\lambda_HH(\pi)+\mathbb{E}_\pi[c(s,a)] \label{eq:rl}
\end{equation}

Specific instances of this process result into apprenticeship learning methods, e.g. the one described in \cite{abbeel2004apprenticeship}, assuming that the cost function is given by a linear combination of basis functions, which result to feature vectors over states and actions. 
The linearity assumption is restrictive for complex problems, such as the trajectory prediction problem in the ATM domain. In addition, the hand-crafted state features are a big engineering burden. Finally, this method is computationally expensive as it runs a Reinforcement Learning algorithm at every cost function update, to find a policy that performs optimally w.r.t. the learnt cost function.

To address linearity limitations and hand-crafted state features, the Guided Cost Learning approach proposed in \cite{finn2016guided} uses neural networks to represent the cost function. It also provides a more computationally efficient approach, by applying a single gradient step for each new update of the cost function, instead of fully optimizing the learned policy in regard to every update of the cost function.
It has been demonstrated in \cite{finn2016connection} that Guided Cost Learning is equivalent to a specific instantiation of the Generative Adversarial Imitation Learning (GAIL) framework \cite{ho2016generative}. 

 GAIL \cite{ho2016generative} can imitate complex behavior as it does not apply restrictive assumptions
on the cost function and scales to large, continuous state-action spaces.  
GAIL directly learns the optimal policy from expert demonstrations, quite efficiently, since it does not need to derive a cost function that will be used by a Reinforcement Learning method to derive a policy. Actually, it aims to bring the distribution of the state-action pairs of the imitator as close as possible to that of the expert.  
GAIL uses an architecture similar to Generative Adversarial Networks to optimize the following objective:
\begin{equation}
    \min_\pi \max_{D\in(0,1)^{SxA}}\mathbb{E}_{\pi}[logD(s,a)]+\mathbb{E}_{\pi_E}[log(1-D(s,a))]-\lambda_H H(\pi) \label{eq:GAIL1}
\end{equation}
where $\pi$ is the imitator policy, $\pi_E$ is the expert's policy, $D$ is a binary classifier called discriminator which distinguishes state-action pairs generated from $\pi$ and $\pi_E$. $H(\pi) \triangleq \mathbb{E}_{\pi}[-log\pi(a|s)]$ is the $\gamma$-discounted causal entropy of the policy $\pi$. As shown in \cite{ho2016generative}, equation (\ref{eq:GAIL1}) provides a way to solve the two steps in imitation process described by equations (\ref{eq:maxent}) and (\ref{eq:rl}). 

To predict aircraft trajectories via imitation learning, we are using the GAIL framework. The GAIL implementation to address the specific problem is described after the trajectory prediction specification problem given in the next Section, in Section \ref{sec:GAIL}.


\section{Problem Specification}
\label{sec:formulation}


Given the abstract specification of the data-driven  trajectory prediction problem  in Section \ref{sec:prediction}, and the formulation of the trajectory imitation learning problem provided in Section \ref{sec:imitation}, we can provide a formulation of the problem we address here: The {\em data-driven aircraft trajectory prediction problem as an imitation learning task}.

Let us assume a set $\textbf{T}_E$= $\{T_{E,i}, i=1,... N\}$ of historical,  enriched aircraft trajectories generated by an expert policy $\pi_E$. These trajectories have various number of states, and therefore, various lengths $|T_{E,i}|$. The objective is to find a policy that minimizes the difference between the expected cumulative cost of the predicted trajectories and of the trajectories in $\textbf{T}_E$, given an approximation of the cost function that penalizes any state-action pair generated by any policy in $\Pi-\{\pi_E\}$. As shown in \cite{ho2016generative}, this objective is equivalent to finding a policy $\pi$ that  brings the distribution of the state-action pairs generated by it, as close as possible to the distribution of the state-action pairs demonstrated by trajectories in $\textbf{T}_E$.

As pointed out in Section \ref{sec:prediction}, in this work we aim to predict the 3D aircraft position at specific time instants, given an initial time instant $t_0$: Specifically, we aim at determining the evolution of the trajectory in space every $\Delta t$ seconds, i.e. at time instances $t_i=t_0+ (\Delta t*i)$, $i=1,2,3...$, given the position of the aircraft at time instance $t_0$.


A crucial decision concerns the set $A$  of actions considered, which should adequately and unambiguously (although, in a non-deterministic way) specify the evolution of the historical as well as of the predicted trajectories. In our approach, and very close to the General Adversarial Imitation from Observations approach described in \cite{torabi2018generative}, we are motivated to focus on states and on their evolution, rather than on the  actual actions that may determine this state evolution. This approach is also motivated by considering the following: (a) Expert trajectories do not specify in any way the actions applied in any state and thus, these have to be determined under specific assumptions that may bring noise into the learning process, (b) there are several possibilities of instruction combinations for evolving the aircraft state, at different levels of detail, which result in a high-dimensional state-action space, and which require considering constraints between instruction combinations, (c) what we aim to actually predict is the evolution of aircraft states in the 4D space (i.e. regarding position and time), and (d) the imitation learning approach that we take aims to bring the distribution of state-action pairs of the imitator close to the corresponding distribution of the expert.

Therefore, we consider that the set $A$ contains all the possible triples ($\Delta l$, $\Delta f$, $\Delta h$) that specify the difference between states' position information in 3D, given the constraint that this difference must be feasible within the constant $\Delta t$ period considered. 
Indeed, these actions can be determined by the demonstrated trajectories unambiguously and very efficiently, although in low-quality surveillance data space-time constraints concerning the evolution of aircraft states may be violated. 
This action set has three additional important effects:
(a) We can tune the resolution of the predicted trajectory by changing the $\Delta t$.
(b)Given a specific $\Delta t$ (e.g. 5 seconds), and the evolution of the trajectory until reaching the destination airport, we can determine the estimated time of arrival (ETA), which is simply ( $\Delta t * |T_\pi|$), given the predicted trajectory $T_\pi$.
(c) The transition between positions is deterministic given an action: Given position $(l,f,h)$  and an action ($\Delta l$, $\Delta f$, $\Delta h$), the  position in the next state is $(l+\Delta l,f+\Delta f,h+\Delta h)$.

Given the above, the {\em data-driven aircraft trajectory prediction problem as an imitation learning task} is specified as follows:

\noindent Given a set $\textbf{T}_E$= $\{T_{E,i}, i=1,... N\}$ of historical,  enriched aircraft trajectories, and a time step  $\Delta t$, we need to determine a policy $\pi \in \Pi$ which optimizes the objective specified by equation (\ref{eq:GAIL1}). This policy, given the initial state of aircraft $s_0=\langle(l_0,f_0,h_0),t_0, v_0\rangle$, determines the evolution of the trajectory at any time instant $t_0+(\Delta t*i)$, $i=1,2,3...$. Specifically, it determines $\pi((\Delta l, \Delta f, \Delta h) | s, \Delta t)$, i.e. the evolution of the aircraft position at state $s_r=\langle(l,f,h),t, v\rangle$ after $\Delta t$ time instants.



\section{Trajectory Prediction Framework}
\label{sec:framework}

This section motivates and provides a description of the overall prediction method, and presents details on the constituent steps.

Generally, given a set of trajectories between any pair of airports, one may detect different patterns of behavior, which may be due to different contextual factors that affect stakeholders' decision making on the evolution of the trajectory. The choice of the runway approaching the destination airport, for instance, may due to airport weather conditions, traffic, or airline preferences, and may result to significantly different trajectories. What we need to do towards automating the data-driven trajectory prediction process is to detect distinct patterns of trajectories, identifying also the features that distinguish between them. Then, we can learn a distinct policy per class of trajectories, i.e. for those trajectories following a specific pattern of behavior.  This can make the learning process much more efficient and effective in contrast to training a single model, considering all possible trajectories. However, to predict a single trajectory we need to know which policy to apply, thus, the mode of behavior it will most probably follow. One solution to this is to forecast the contextual features that may impact the evolution of the trajectory and determine the class of the future trajectory using these features. This classification step is thus restricted to those features, which they do distinguish between different modes of behavior, and can be forecast or can be known at any stage (tactical or pre-tactical) of ATM operations. 

Thus, the trajectory prediction approach that we propose incorporates a trajectories  clustering step, a classification step, and finally a trajectory imitation step, solving the following refinement of the above-specified data-driven trajectory prediction problem: 

Given (a) the set $\textbf{T}_E$= $\{T_{E,i}, i=1,... N\}$ of demonstrated trajectories, (b) a time step  $\Delta t$, $K$ distinct classes $C_{E,l}, l=1,2...K$ of $\textbf{T}_E$, and (c) a set $\textbf{S}^f$ of states $s^{f}_{{fix}_r}=\langle(l_{fix},f_{fix},h_{fix}),t^f, v^f\rangle$ at specific ``landmark" (fixed) positions enriched with forecast contextual features, where $t^f$ denotes the forecast time instance that the trajectory will reach the position ($l_{fix},f_{fix},h_{fix}$), and $v^f$ denotes a vector of forecast contextual features at that point at time $t^f$, we aim to determine 
(i) the specific class $C_{E,\sigma} \subseteq \textbf{T}_E$ of expert trajectories that most probably the future trajectory crossing the points in $\textbf{S}^f$ belongs, and (ii) 
a policy 
$\pi_\sigma \in \Pi$ which optimizes the objective specified by equation (\ref{eq:GAIL1}) given the demonstrated trajectories in the determined class $C_{E,\sigma}$, and which specifies the evolution of the trajectory at any time instant $t_0+(\Delta t*i)$, $i=1,2,3...$ given the initial state of aircraft $s_0=\langle(l_0,f_0,h_0),t_0, v_0\rangle$.

The subsequent subsections describe the methods used for each of the different steps.


\subsection{Trajectory Clustering}\label{clustering}
\label{sec:clustering}

First we aim to divide $\textbf{T}_E$ into a set of $K$  clusters $C_{E,l}$, $l=1...K$ in such a way that trajectories belonging to the same cluster represent a pattern of behavior that is more similar compared to the behavior of trajectories outside this group \cite{pattern_recognition}. Generally, it holds that $\cup_{l=1}^KC_{E,l} \subseteq \textbf{T}_E$, considering that outliers may not be assigned in any cluster, and $C_{E,i} \cap C_{E,j} = \emptyset \text{, for } i \neq j$.

Towards this goal we apply an agglomerative clustering which is a bottom-up hierarchical strategy initially treating each trajectory as an individual cluster (singleton) and iteratively merges similar clusters stopping when only K clusters remain. To merge two clusters we use the average distance among all their member trajectories, using the Ward merging criterion \cite{doi:10.1080/01621459.1963.10500845}. 

The distance measure we use is the normalized Dynamic Time Warping (DTW) measure \cite{pazzani}, which in our case is applied to (a) find optimal matches between two trajectories, and (b) measure the trajectories'  similarity without considering their variable lengths and the variable time distances between subsequent points. In our case the following DTW formulation has been applied:
\begin{equation}\label{nDTW}
nDTW(T_i,T_j)={DTW(T_i,T_j)}/{\sqrt{d*max(|T_i|,|T_j|)}}
\end{equation}
where $d$ is the number of dimensions in the multivariate data observed. The denominator can be seen as the largest DTW distance between two trajectories, thus bounding nDTW in [0,1]. 

Considering that the appropriate number $K$ of clusters is unknown, 
the problem of determining $K$  can be transferred to a silhouette coefficient maximization problem \cite{de_amorrin}. The computation of the silhouette coefficient needs only pairwise distances and the calculation of clusters' centroids is avoided. 

\subsection{Future Trajectory Classification}\label{classification}
\label{sec:classification}


The future trajectory classification problem we consider in our case is as follows: Given a set of $K$  clusters $C_{E,l}$, $l=1...K$, and the set $\textbf{S}^f$ of forecast enriched states at specific ``landmark" positions that the future trajectory will cross, we aim to determine 
the class $C_{E,\sigma}$ of demonstrated trajectories that most probably the trajectory that will cross the enriched points in $\textbf{S}^f$ belongs. 

These fixed positions may be waypoints (fixes) declared in a planed trajectory (e.g. the flight plan), although in this article we consider a single point that the trajectory will cross for sure: The destination airport. 
Specifically, we consider the singleton $\textbf{S}^f= \{\langle(l_{dest},f_{dest},h_{dest})$, $t^f, v^f_{dest}\rangle\}$, with an enriched state corresponding to the destination airport, reached at $t^f$, which is equal to the estimated time of arrival. The $v^f_{dest}$ comprises destination airport's forecast metereological variables (specified in Section \ref{sec:experiments}), although more features can be incorporated (e.g. traffic conditions).

The classifier used in the pipeline is the random forest classification algorithm\cite{rf}, which is trained with enriched trajectories in $\textbf{T}_E$ being assigned to the specific clusters identified, and is called to predict to which cluster the future trajectory crossing $\textbf{S}^f$ most probably belongs. It must be noted that each training trajectory is enriched with all the variables corresponding to those in  $v^f_{dest}$, being assigned with the real (not forecast) values at the time of flight arrival in $(l_{dest},f_{dest},h_{dest})$.

\subsection{GAIL: Learning to imitate trajectories}\label{bc_gail}
\label{sec:GAIL}

As specified above, we are using GAIL for imitation learning. Actually GAIL is trained per cluster, thus revealing a policy $\pi$ per cluster. Simply, GAIL employs a generative trajectory model $G$ that models $\pi$ and a discriminative classifier $D$ that distinguishes between the distribution of data (i.e. state action pairs)  generated by the policy and the demonstrated data. Both $\pi$ and $D$ are represented by function approximators, with weights $\theta$ and $w$, respectively. Following the implementation described in \cite{ho2016generative}, GAIL alternates between an Adam \cite{adam} gradient step on $w$ to increase equation (\ref{eq:GAIL1}) with respect to $D$, and a step on $\theta$ using the Trust Region Policy Optimization (TRPO) algorithm \cite{schulman2015trust} to decrease equation (\ref{eq:GAIL1}). TRPO optimizes the following objective:
\begin{equation}
\begin{aligned}
 \underset{\theta}{minimize} \mathbb{E}_{s\sim\rho_{\theta_{old}},a\sim q}[\frac{\pi_{\theta}(a|s)}{q(a|s)}Q_{\theta_{old}}(a|s)]\\
subject\: to\: \mathbb{E}_{s\sim\rho_{\theta old}}[D_{KL}(\pi_{\theta_{old}}(\cdot |s)\| \pi_{\theta}(\cdot|s))] \leq \delta
\end{aligned}
\end{equation}
where $\rho_{\theta_{old}}$ is the distribution of states generated using the prior-to-update (old) policy $\pi_{\theta_{old}}$, $q$ is an action sampling distribution that we consider equal to $\pi_{\theta_{old}}$, $\pi_{\theta}$ is the updated policy with parameters $\theta$, $Q_{\theta_{old}}$ is the state-action value function of the old policy and $\delta$ is a constant that constraints the KL divergence between $\pi_{\theta_{old}}$ and $\pi_{\theta}$, preventing the policy from changing too much due to noise in the policy gradient.

We approximately solve the TRPO optimization problem as described in \cite{schulman2015trust} Appendix C, using the conjugate gradient method and a line search. In our setting we set $\lambda_H=0$, so we omit $-\lambda_HH(\pi)$ from the equation (\ref{eq:GAIL1}), following the practice demonstrated in \cite{ho2016generative}.

A  subtle point in our implementation is that, instead of approximating $Q$, we utilize a separate critic model to approximate the state advantage defined as $A_t=A(s_t,a_t|\pi)=Q(s_t,\pi(s_t))-V^{\pi}(s_t)$, aiming to lower the gradient variance. We follow the Generalized Advantage Estimation (GAE) approach introduced in \cite{gae}, which provides a balance between low variance and a small amount of bias introduced. Formally, we estimate the advantage from the sampled state-action pairs as follows:
\begin{equation}
\begin{aligned}
\hat{A}^{GAE(\gamma,\lambda)}_t\coloneqq(1-\lambda)(\hat{A}^{(1)}_t+\lambda\hat{A}^{(2)}_t+\lambda^2\hat{A}^{(3)}_t+\dots)
\end{aligned}
\end{equation}
where $\gamma \in [0,1]$ is the discounting factor, $\lambda$ a hyper-parameter and 
\begin{equation}
\begin{aligned}
\hat{A}^{(k)}_t\coloneqq-V(s_t)+r_t+\gamma r_{t+1}+\dots+\gamma^{k-1}r_{t+k-1}+\gamma^{k}V(s_{t+k})
\end{aligned}
\end{equation}

Algorithm \ref{alg:GAIL}  shows the aforementioned procedure in more detail. Specifically, we pre-train $G$ using Behavioral Cloning. Then, at each GAIL iteration, the algorithm samples from the initial state distribution and  generates roll-out trajectories. It uses the generated state-action samples and the samples of the historical trajectories to update the $D$ parameters $w$. $D$ is updated with cross entropy loss that pushes the output for the demonstrated state-action samples closer to 0 and $\pi_{\theta}$ state-action samples closer to 1. Next, the imitation algorithm takes a policy step using the TRPO \cite{schulman2015trust} update rule and $logD(s,a)$ as the cost function approximation to update $\theta$.  It must be noted that the   $t$ parameter in the denotation of the approximation of the state advantage in Algorithm \ref{alg:GAIL} is left implicit, for simplicity of the presentation.

\begin{algorithm}
\small
\caption{GAIL} \label{alg:GAIL}
\begin{algorithmic}[1]
\STATE {\textbf{Input:} Expert trajectories $\tau_{E}\sim\pi_{E}$, initial policy $\pi_{\theta_0}$ and discriminator parameters $w_0$}
\STATE {\textbf{Output:} Policy $\pi_{\theta}$}
\STATE {Initialize policy using Behavioral Cloning.}
\FOR{i=0,1,2,...}
    \STATE{Sample trajectories $\tau_i\sim\pi_{\theta_i}$}
    \STATE{Update $D$ parameters $w$ with the gradient}
    \STATE{$\hat{\mathbb{E}}_{\tau_{i}}[\nabla_w log(D_w(s,a))] + \hat{\mathbb{E}}_{\tau_{E}}[\nabla_w log(1-D_w(s,a))]$}
    \STATE{Estimate advantages $\hat{A}^{GAE(\gamma,\lambda)}$, according to $\pi_{\theta_{old}}$}
    \\
\STATE{Take a policy step using the TRPO rule with cost function $log(D_w(s,a))$:}
    \\
        Take a KL-constrained natural gradient step with
        $\hat{\mathbb{E}}_{\tau_{i}}[\nabla_\theta     
    \frac{\pi_{\theta}(a|s)}{\pi_{\theta_{old}}(a|s)}\hat{A}^{GAE(\gamma,\lambda)}]$
        subject to
       $\mathbb{E}_{s\sim\rho_{\theta old}}[D_{KL}(\pi_{\theta_{old}}(\cdot |s)\| \pi_{\theta}(\cdot|s))] \leq \delta$
\STATE{\textbf{end for}}
\ENDFOR
\end{algorithmic}
\end{algorithm}
\section{Experimental Evaluation}
\label{sec:experiments}

\subsection{Experimental Setting}\label{exp_settings}
Datasets exploited in our experiments include radar tracks (surveillance data representing raw trajectories) for flights from Barcelona to Madrid and from the \nth{1} to the \nth{24} of April 2016 , weather data obtained from National Oceanic and Atmospheric Administration (NOAA), and weather reports from airports (METAR). The aim is to predict  trajectories for this origin-destination pair.

Given these datasets, demonstrated trajectories are enriched with eleven (11) numerical variables  corresponding to  (a) 6 meteorological features at the corresponding 3D state position and time, provided by NOAA, and 5 features specifying actual weather conditions at the arrival airport at the time of arrival, provided by METAR.  The NOAA features are \textit{pressure surface, relative humidity isobaric, temperature isobaric, wind speed gust surface, u-component of wind isobaric, v-component of wind isobaric}. Features from METAR include \textit{wind direction, wind speed in knots, pressure altimeter in inches, visibility in miles, wind gust in knots}.

The set of raw trajectories has been pre-prosessed and cleaned. The pre-processing stage interpolates points in trajectories, so that two points have a temporal distance of $\Delta t=5$ seconds.
 This task calculates the average velocity of the aircraft between subsequent points in the original  trajectory. Assuming a constant velocity between these points we can calculate the position of the aircraft every $\Delta t$ seconds, and we finally reconstruct the trajectory keeping only the points occurring every $\Delta t$ seconds along the original trajectory. This  is important in order to exclude the temporal dimension from actions.
The cleaning task aims to detect incomplete trajectories starting or finishing away from any of the airports, as well as flights  showing inconsistent behavior (e.g. covering a significant distance within an unreasonably small amount of time), due to imperfections in the raw data.

The resulting set of 528 trajectories from Barcelona to Madrid has been randomly divided into a set $\textbf{T}_E$ of 478 trajectories and a test set of 50 trajectories. However, the clustering algorithm clusters all 528 trajectories. Doing so, we are able to measure the accuracy of the trajectory classification algorithm, in conjunction to the accuracy of the trajectory imitation process. 

\begin{figure}%
\centering
      \includegraphics[width=0.85\columnwidth]{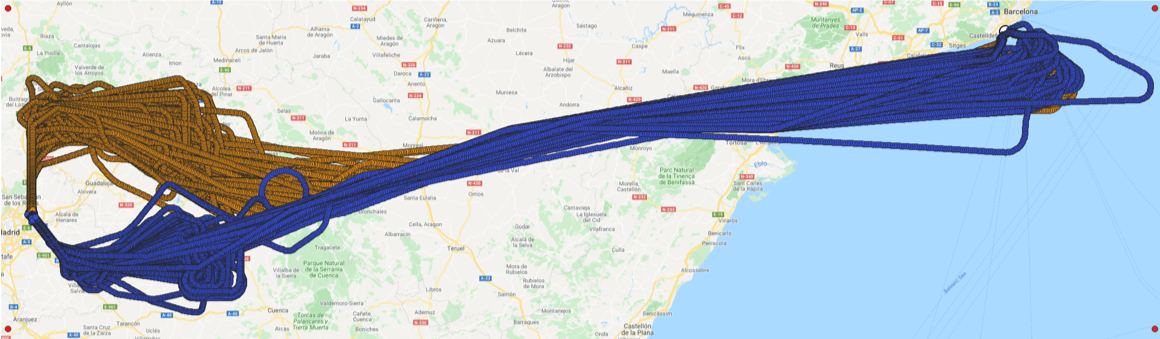}
\caption{Clusters of trajectories from Barcelona to Madrid (visualised with QGIS).}
\label{fig:clusters_qgis}
\end{figure}

The clustering process resulted into $K=2$ clusters each with 250 and 278 trajectories, taking into account all the features in the enriched trajectories. Each cluster shows a different pattern of approaching the Madrid airport, as depicted in Figure \ref{fig:clusters_qgis}. Then, considering only the $\textbf{T}_E$ trajectories (i.e. after excluding the 50 test trajectories), GAIL was trained for each of the two clusters, providing two policies corresponding to the distinct behavioral modes. Subsequently, we also provide results with a single policy, after training GAIL in $\textbf{T}_E$, without considering clusters, and thus avoiding the future trajectory classification task.

During testing, in order to determine the policy to be used for prediction, we followed the classification approach described in Section \ref{sec:classification}, considering $\textbf{S}^f= \{\langle(l_{MAD},f_{MAD},h_{MAD}),t^f, v^f_{MAD}\rangle\}$, with an enriched state corresponding to the destination airport, and $t^f$ estimated by considering the average duration of the Madrid-Barcelona flights in $\textbf{T}_E$. The features vector $v^f_{MAD}$ comprises destination airport's {\em forecast} meteorological variables corresponding to the five METAR variables mentioned above. The classifier was trained using 5-fold cross validation that has been repeated for all combinations of hyperparameters. This process resulted in a classification method that samples trajectories with replacement, with 20 trees of max depth equal to 20, with leaf nodes comprising at least one trajectory, and with at least two trajectories in a node before  splitting it.

To implement the generative model $G$ and the discriminator $D$ in GAIL we have used two neural networks, each  consisting of two dense layers of 100 nodes, each layer with $tanh$ activation. The input  for $G$ corresponds to the four 3D position and temporal variables per state, and the six meteorological features provided by NOAA. $D$   takes as additional input the three action variables. $G$ has a dense output layer with size equal to the number of action variables (i.e. three), while the  output layer of $D$ has one node.  $G$ outputs for each action variable the mean of a Gaussian distribution with logarithm of standard deviation equal to 0.9, resulting to a stochastic  policy. To initialize the policy's parameters, we use Behavioral Cloning minimizing the Mean Square Error between demonstrated actions and the policy actions, over the training set, using Adam optimization. This has been trained with 100 epochs and  10 fold cross validation.
GAIL is trained for 1500 batches. At each round the policy generates a batch of 50000 state-action samples. The number of episodes needed to acquire this number of samples is not constant. At each episode the method randomly selects a starting point regarding a trajectory in the training set and uses $G$ to generate roll-outs.  Roll-outs terminate either when a trajectory point lies within a 5km radius from the destination airport, or when the trajectory has 1000 points, or when it lies outside the bounding box defined by the geographic (lon,lat) coordinates $(-3.7038, 41.4)$ , $(2.9504, 39.9864)$ corresponding to the red dots in the corners of Figure \ref{fig:clusters_qgis}. These 50000 samples are used for training the Discriminator $D$. Specifically, we use Adam optimization and 100 epochs to maximize equation (\ref{eq:GAIL1}) w.r.t. the $D$ parameters $w$. 

To evaluate the proposed approach we provide results regarding the prediction of Barcelona-Madrid trajectories, in the following experimental settings: (a) Using the prediction pipeline and two policies modelling the behavioral modes shown by the two clusters identified (denoted as ''MultPolicies" approach), (b) using one policy modelling the behavior of all flights (denoted as ``OnePolicy'' approach)\footnote{The OnePolicy approach uses a $G$ neural network whose input is extended to include the five METAR variables used in the classification stage of the MultPolicies approach, evaluating its ability to distinguish between the detected behavioral modes.}. For each of the settings, and in order to show the prediction abilities of the proposed method, we evaluate the prediction accuracy in 5 cases, by considering as initial state, a state after ($M*flight\_duration$) minutes from $t_0$, where $M \in \{0, 0.2, 0.5,0.7\}$: Results show the capacity of the method either at the pre-tactical (i.e. for $M=0$) or at any state during the tactical stage of operations.

The results reported are generated from 20 independent experiments per setting, considering each of the 50 test trajectories, resulting in aggregating results from 1000 experiments per setting/case combination.  Specifically, we report on the trajectory  prediction accuracy using the following measures: (a) Root Mean Square Error (RMSE) in meters in each of the 3 dimensions, as well as in 3D, (b) Along-Track Error (ATE), (c) Cross-Track Error (CTE), and (d) Vertical deviation (V). ATE and  CTE are according to the methodology proposed in \cite{doi:10.2514/6.2004-4788}. The along track error is measured parallel to the predicted trajectory,  while the cross track error is measured perpendicular to the predicted course. All measures are computed for each predicted trajectory point after computing its corresponding point in the test trajectory using the DTW method. Trajectories are not segmented.  Finally, we  provide results on the estimated time of arrival (ETA) according to predictions, compared to the arrival time of test trajectories.  

\subsection{Results}
\newcommand\mc{\multicolumn}
Before delving into the results provided by the imitation learning method, we need to point out that the average accuracy of 100 independent experiments of the future trajectory classifier is 0.976, with a standard deviation of 0.0094. This proves the fact that the classification method is suitable, but more importantly, that the destination airport’s meteorological forecast variables are important. More fixed points and/or features can be added in future enhancements.

\begin{table}
\centering
\footnotesize
\caption{RMSE (meters) results for all cases.}
\label{tab:RMSE}
\begin{tabular}[]{|ll|*8{c|}}
  \cline{3-10}
  \mc{1}{l}{}&&\mc{4}{|c|}{OnePolicy}&\mc{4}{c|}{MultPolicies}\\
  \cline{3-10}
  \mc{1}{l}{}&& \mc{1}{c}{Long} & \mc{1}{c}{Lat} & \mc{1}{c}{Alt} & \mc{1}{c|}{3D} 
              & \mc{1}{c}{Long} & \mc{1}{c}{Lat} & \mc{1}{c}{Alt} & \mc{1}{c|}{3D} \\
  \hline
    & 0 & 14350 & 8347 & 457 & 17279 & 10932 & 5577 & 333 & 12652\\
  \cline{3-10}
    & 0.2 & 13780 & 8311 & 550 & 16825 & 10252 & 5477 & 402 & 12048\\
  \cline{3-10}
    & 0.5 & 9726 & 8847 & 427 & 14066 & 7490 & 6679 & 324 & 10565\\
  \cline{3-10}
    & 0.7 & 5979 & 7059 & 246 & 9916 & 4430 & 6360 & 188 & 8033\\
  \cline{3-10}
  \hline
\end{tabular}
\end{table}

\begin{table}
\centering
\footnotesize
\caption{ATE, CTE, \& V (in meters), and ETA Error (in seconds) for all cases.}
\label{tab:ACTE}
\begin{tabular}[]{|ll|*8{c|}}
\cline{3-10}
  \mc{1}{l}{}&&\mc{4}{|c|}{OnePolicy}&\mc{4}{c|}{MultPolicies}\\
  \cline{3-10}
  \mc{1}{r}{}&& \mc{1}{c}{ATE} & \mc{1}{c}{CTE}  & \mc{1}{c|}{V} & \mc{1}{c|}{ETA}
              & \mc{1}{c}{ATE} & \mc{1}{c}{CTE}  & \mc{1}{c|}{V} & \mc{1}{c|}{ETA}\\
  \hline
    & 0 & -31.6 & 577.0 & 67.0 & 245.96 & 305.8 & 154.0 & 23.8 & 274.10 \\
  \cline{3-10}
    & 0.2 & -99.4 & 808.5 & 121.3 & 288.65 & 454.5 & 391.8 & 57.6 & 268.70 \\
  \cline{3-10}
    & 0.5 & 984.9 & 1657.1 & 115.6 & 398.34 & 826.0 & 875.1 & 79.5 & 325.84\\
  \cline{3-10}
    & 0.7 & 851.5 & 1540.7 & 25.8 & 460.56 & 1065.0 & 1133.0 & 5.1 & 369.03\\
  \cline{3-10}
  \hline
\end{tabular}
\end{table}

Table \ref{tab:RMSE} shows the mean RMSE error of the predicted vs the actual (test) trajectory  in meters for each of the three dimensions and in 3D; while Table \ref{tab:ACTE} shows the mean ATE, mean CTE, as well as the mean V, in meters.  It also reports the mean error of the expected arrival time (ETA) in seconds for each case. Both tables are split to  the OnePolicy and the MultPolicies settings results, while the rows correspond to the different values of $M$.

Figures in Table \ref{tab:box_plots} show box plots for all the measures. The y axis specifies the error measured. Horizontal lines of each box plot represent the \nth{25}, the \nth{50}, the \nth{75} and the \nth{100} percentile. Diamonds indicate outliers and the numbers indicate the medians. The left column provides RMSE and the right the track errors. Again, the rows correspond to the different values of $M$.

Regarding the results, it must be noted that the MultPolicies setting provides consistently better results compared to the One Policy setting, thus providing evidence on the efficacy of the proposed pipeline. This is due to the fact that the OnePolicy setting fails to model effectively the different behavioral modes, predicting trajectories that in the worst case follow a different pattern of behaviour than the actual one. Low deviations of predicted from the actual trajectories, compared to state of the art methods (Section \ref{sec:related}) provide firm evidence of the  imitation learning approach efficacy.

Table 1 shows that the proposed method is quite effective to predict the whole trajectory at the pre-tactical stage ($M=0$), while the RMSE is reduced while increasing $M$, i.e. while we select a starting point far from the origin airport, simulating the tactical stage. However, Table 2 shows that the mean along and cross track errors increase while increasing $M$, which is most probably due to the complexities of the trajectories while approaching the destination  airport (i.e. due to holding patterns, maneuvers, etc.). Thus, it seems that a more refined approach must be used to address the landing part of the trajectory more accurately.

\begin{table*}[ht]
\small
\caption{RMSE and Track Errors in meters for all cases.}
\label{tab:box_plots}
\centering
\begin{tabular}{l | c  c}
    $M=0$
    & \includegraphics[width=0.4\textwidth]{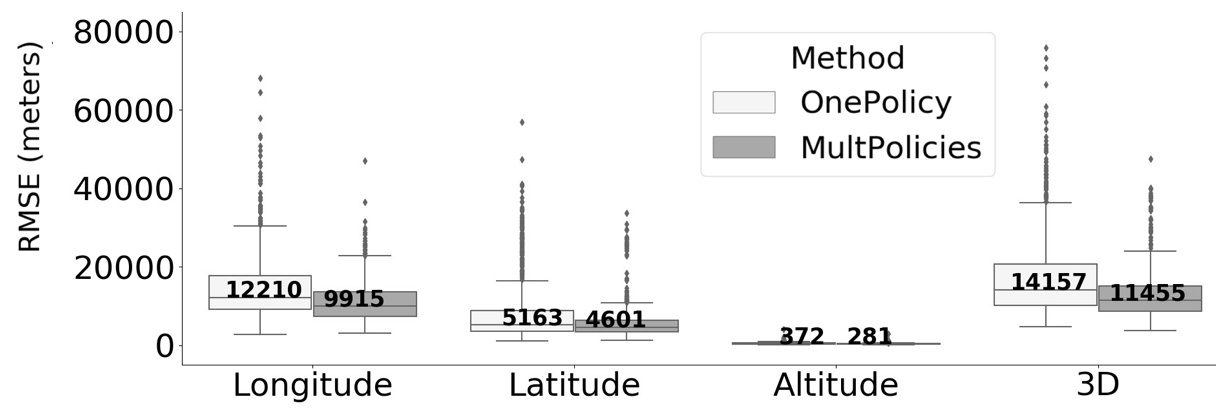}
    & \includegraphics[width=0.5\textwidth]{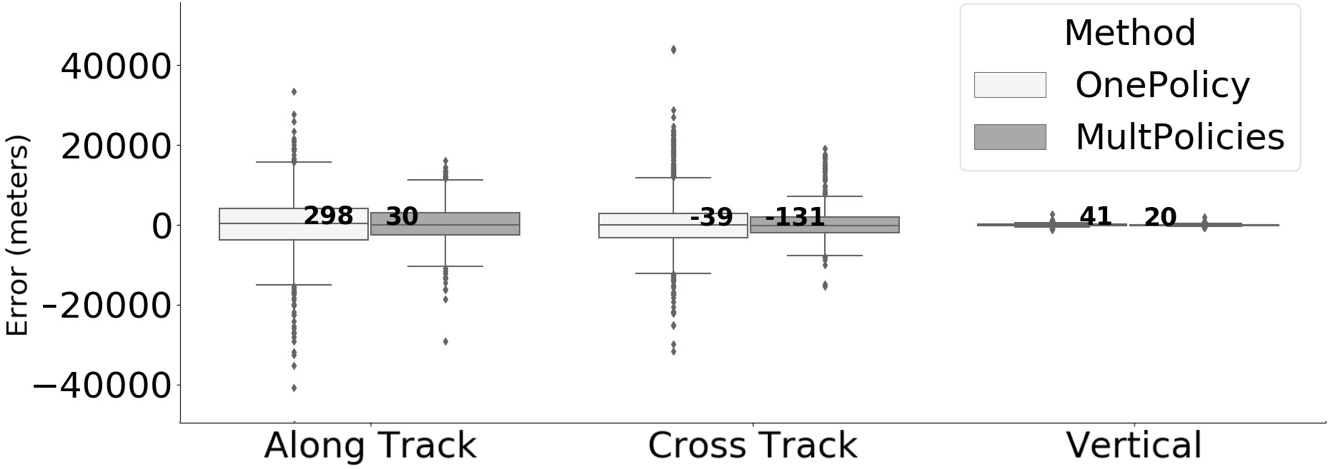} \\
\hline
    $M=0.2$
    & \includegraphics[width=0.4\textwidth]{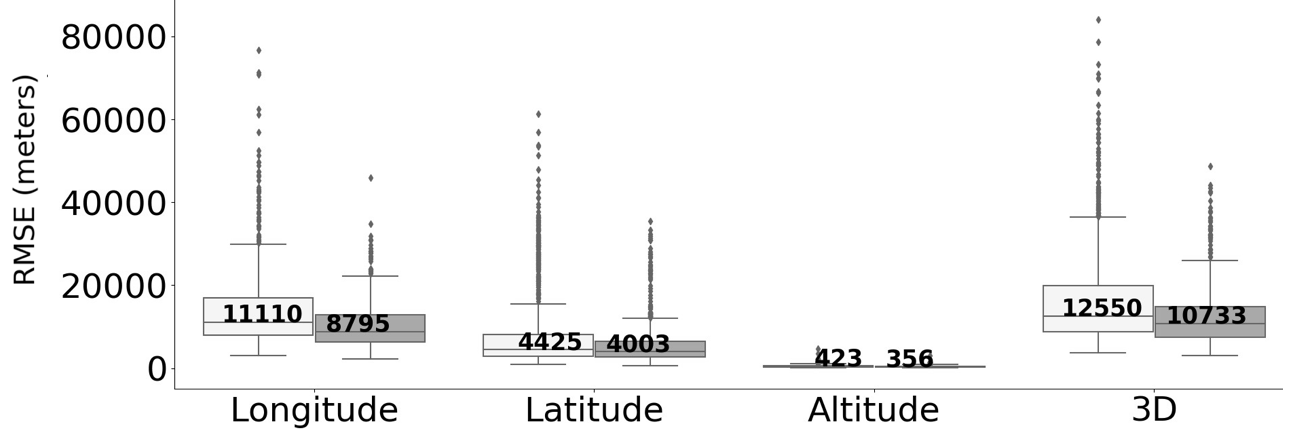}
    & \includegraphics[width=0.5\textwidth]{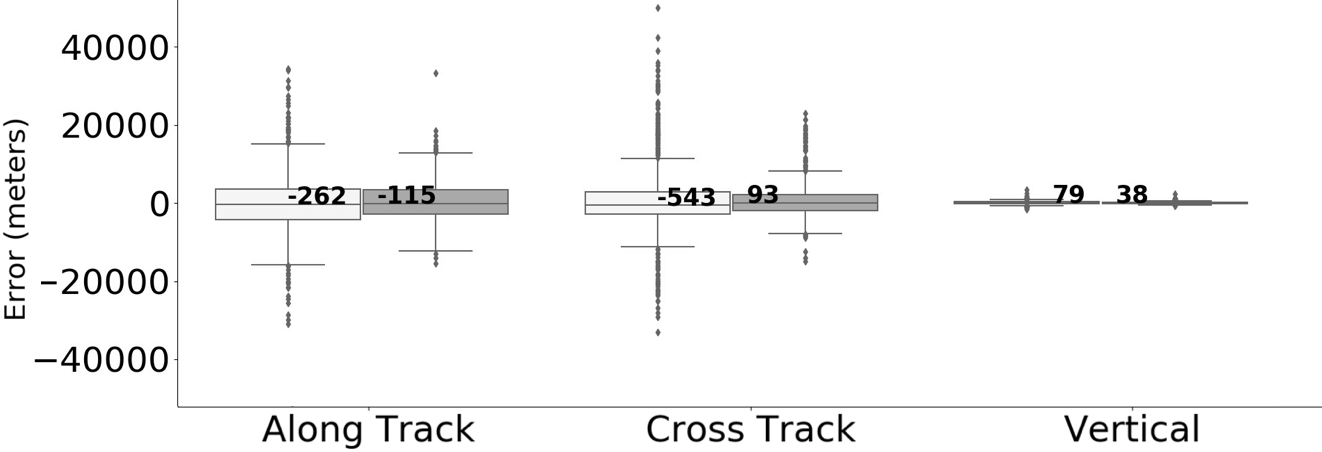} \\
\hline
    $M=0.5$
    & \includegraphics[width=0.4\textwidth]{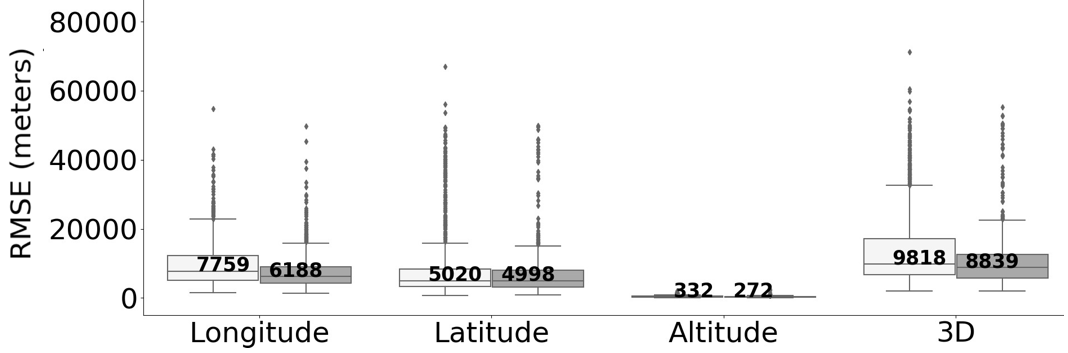}
    & \includegraphics[width=0.5\textwidth]{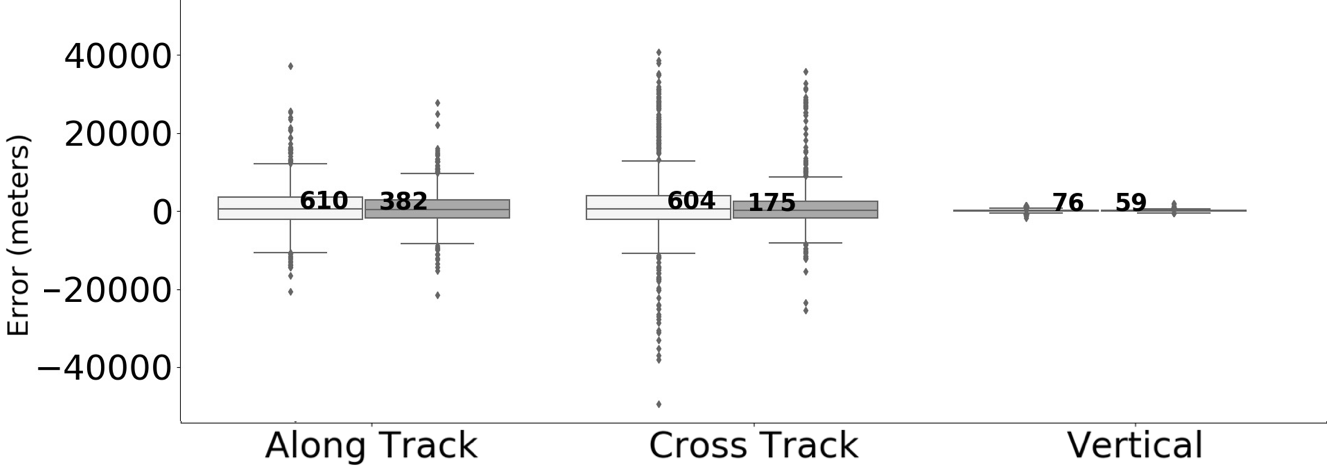} \\
\hline
    $M=0.7$
    & \includegraphics[width=0.4\textwidth]{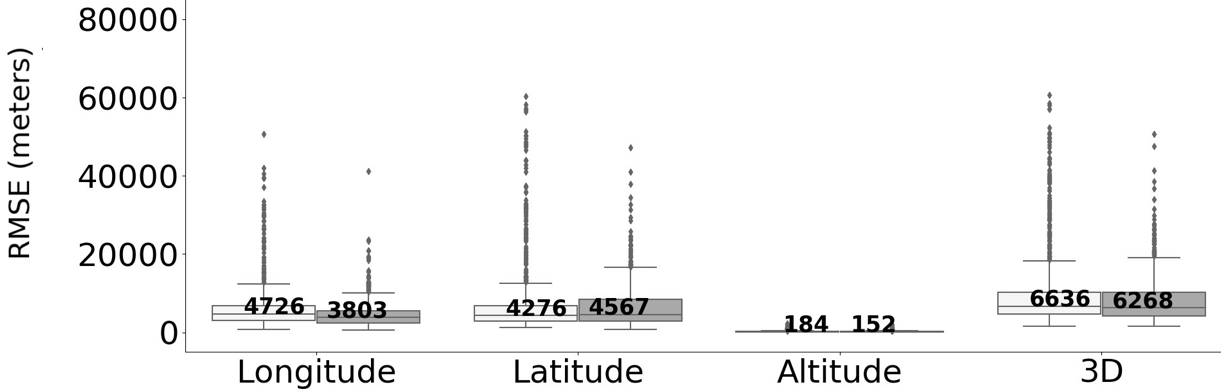}
    & \includegraphics[width=0.5\textwidth]{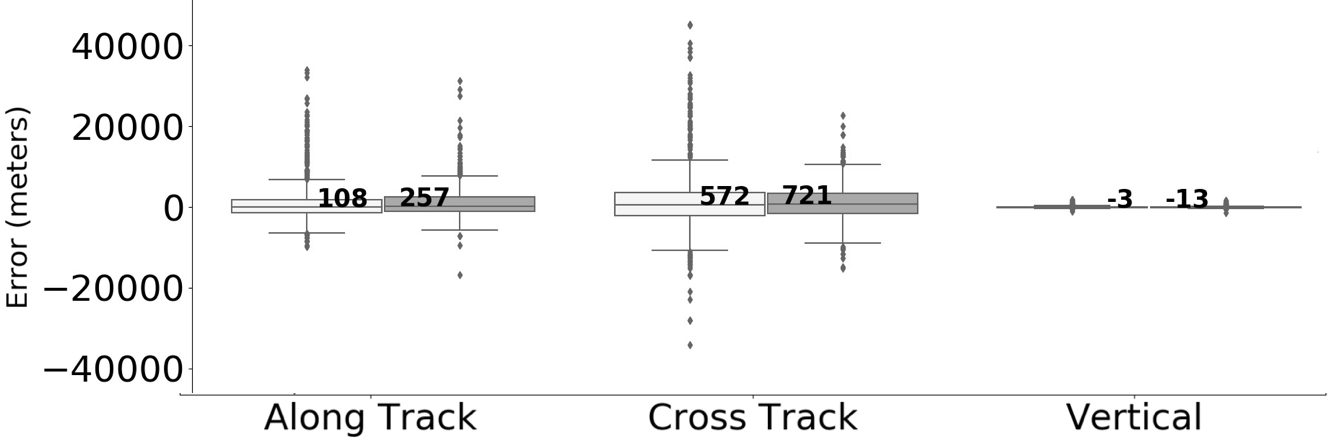} \\
\hline
\end{tabular}
\end{table*}









\section{Related Work}
\label{sec:related}

Reinforcement learning techniques inherently deal with trajectories, formed as policies in an action-state space. Such methods have been used in predicting aircraft trajectories \cite{BRTE}, as well as human and vehicle trajectories in urban spaces  with traffic/crowd. 
The DART \cite{DART} reinforcement learning approach for aircraft trajectory prediction exploits historical trajectories enriched with Aircraft Intent information. The action set in this case includes commands executed by the aircraft Flight Management System. This approach needs a model-based trajectory prediction method in the loop to predict the next aircraft position given a set of commands, incurring a significant computational cost in the whole process, while it requires discretization of state-action parameters, and learning “constraints” on the valid combinations of commands. 
As far as we know, our approach is the first to apply deep imitation learning methods to predict trajectories in the aviation domain.

As pointed out in the introduction, recent data-driven efforts in the field of aircraft trajectory prediction have explored the application of statistical analysis and machine learning techniques. A comprehensive review of trajectory prediction methods in different domains can be found in  \cite{georgiou2018moving}. As far as aircraft trajectory prediction is concerned, most approaches make specific assumptions concerning the types of aircraft considered (e.g. \cite{doi:10.2514/6.2013-4782}, the operational phase considered (e.g. climbing, being in terminal airspace, etc.) (e.g. \cite{Hamed2013StatisticalPO}, \cite{Yang2015TerminalAreaAI}), the look-ahead time ( as in \cite{Hamed2013StatisticalPO} and \cite{Cheng2003DataMF}), or consider specific constraints for making predictions \cite{georgiou2018moving}. 
State of the art approaches in the ATM domain that are closely related to our work are those in \cite{10.1145/2939672.2939694}, \cite{liu2018predicting} and \cite{georgiou2018moving}.

Authors in \cite{10.1145/2939672.2939694} introduce a novel stochastic approach,
modeling trajectories in space and time by using a set of
spatiotemporal 4D joint data cubes, enriching these with aircraft motion parameters and weather conditions. This approach computes the most likely
sequence of states derived by a Hidden Markov Model (HMM), which has been
trained over enriched with weather variables trajectories. The algorithm computes the maximal probability of the optimal state sequence, which is best aligned
with the observation sequence of the aircraft trajectory.  Given that the lateral resolution of each cube is 13km and temporal resolution is 1hr, authors  conclude
that the mean value for the cross-track error (12.601km when the sign is omitted or -3.444km when signed)  is within the boundaries of the spatial resolution. However, our proposed method provides a much lower error along and cross track, with a very low vertical error compared to the 687.497 ft reported there, without limiting the resolution of trajectories' representation, while learning/predicting in continuous action-state space. 

Compared to \cite{liu2018predicting}, the method proposed here is much more effective in terms of predicted trajectory deviations from the actual trajectories in all dimensions, given also that, that approach requires flight plans, as well as a number of actual trajectory points prior to prediction. Authors propose a tree-based matching algorithm to construct image-like feature maps from high-fidelity meteorological datasets. They then model the   trajectory points as conditional Gaussian mixtures with parameters to be learned from the proposed deep generative model, which is an end-to-end convolutional recurrent neural network that consists of a long short-term memory (LSTM) encoder network and a mixture density LSTM decoder network. 

Finally, the approach in \cite{georgiou2018moving} is a ``constrained'' approach, learning the deviations of trajectories from flight plans and reporting low deviations per waypoint. This is in contrast to the  proposed approach, which does not exploit any information constraining the predicted trajectory, although it is generic enough to incorporate such constraints by means of forecast states. The effectiveness of incorporating such constraints in the prediction process is within our future plans.

\section{Conclusions and Future Work}
In this paper we specify the data-driven trajectory prediction problem as an imitation learning task. Towards solving this problem we present a comprehensive framework comprising the Generative Adversarial Imitation Learning state of the art method, in a pipeline with trajectory clustering and classification methods. Evaluation results show the effectiveness of the method to make accurate predictions for the whole trajectory (i.e. with a prediction horizon until reaching the destination) both at the pre-tactical (i.e. starting at the departure airport at a specific time instant) and at the tactical (i.e. from any state while flying) stages, compared to state of the art approaches.
Future Plans include (a) verifying the effectiveness of the method for different origin-destination airports, (b) exploit flight plans to constrain the prediction pipeline, (c) trying to generalize beyond specific origin-destination pairs.
\\ \\
\small 
\textbf{Acknowledgements}: This research is being supported by ENGAGE KTN Catalyst and PhD projects, and partially by Greek National Funds for datAcron for the year 2018. We would like to thank our colleagues C. Spatharis and K. Blekas who implemented the clustering algorithms, G.M. Santipantakis who enriched the raw trajectories, as well our partners Boeing Research and Technology Europe and CRIDA for providing datasets.

\end{document}